\documentclass{sig-alternate}
\pdfoutput=1

\usepackage{amsmath,amsfonts}
\usepackage[binary-units=true]{siunitx}
\usepackage{graphicx}
\usepackage{subfig}
\usepackage{url}
\usepackage{xspace}
\usepackage{tikz}
\usepackage{booktabs}
\usepackage{natbib}
\usepackage[noconfig]{refstyle}
\input{refconfig.cfg}
\usepackage[ruled, linesnumbered]{algorithm2e}
\usepackage{fixme}
\fxsetup{
    status=draft,
    layout=inline,
    theme=color
}

\renewcommand\vec[1]{\mathbf{#1}}

\DeclareMathOperator*{\argmax}{\arg\,\max}

\newcommand{\kmmm}{{\mbox{k-means{-}{-}}}}
\newcommand{\kmeansalgo}{{\ensuremath{k}-means}}
\newcommand{\klmeansalgo}{{\ensuremath{k}-means{-}{-}}}

\begin{document}


\title{Integer Programming Relaxations for Integrated Clustering and
Outlier Detection}

\numberofauthors{1}
\author{
    \alignauthor
    Lionel Ott, Linsey Pang, Fabio Ramos, David Howe, Sanjay Chawla\\
        \affaddr{School of Information Technologies}\\
        \affaddr{University of Sydney} \\
        \affaddr{Sydney, Australia}\\
        \email{lott4241@uni.sydney.edu.au}
}

\maketitle

\begin{abstract}
In this paper we present methods for exemplar based clustering with
outlier selection based on the facility location formulation. Given a
distance function and the number of outliers to be found, the methods
automatically determine the number of clusters and outliers. We
formulate the problem as an integer program to which we present
relaxations that allow for solutions that scale to large data sets. The
advantages of combining clustering and outlier selection include: (i)
the resulting clusters tend to be compact and semantically coherent (ii)
the clusters are more robust against data perturbations and (iii) the
outliers are contextualised by the clusters and more interpretable, i.e.
it is easier to distinguish between outliers which are the result of
data errors from those that may be indicative of a new pattern emergent
in the data. We present and contrast three relaxations to the integer
program formulation: (i) a linear programming formulation (LP) (ii) an
extension of affinity propagation to outlier detection (APOC) and (iii)
a Lagrangian duality based formulation (LD). Evaluation on synthetic as
well as real data shows the quality and scalability of these different
methods.
\end{abstract}



\section{Introduction}
\label{sec:introduction}

\begin{figure}[bt]
    \centering

    \includegraphics{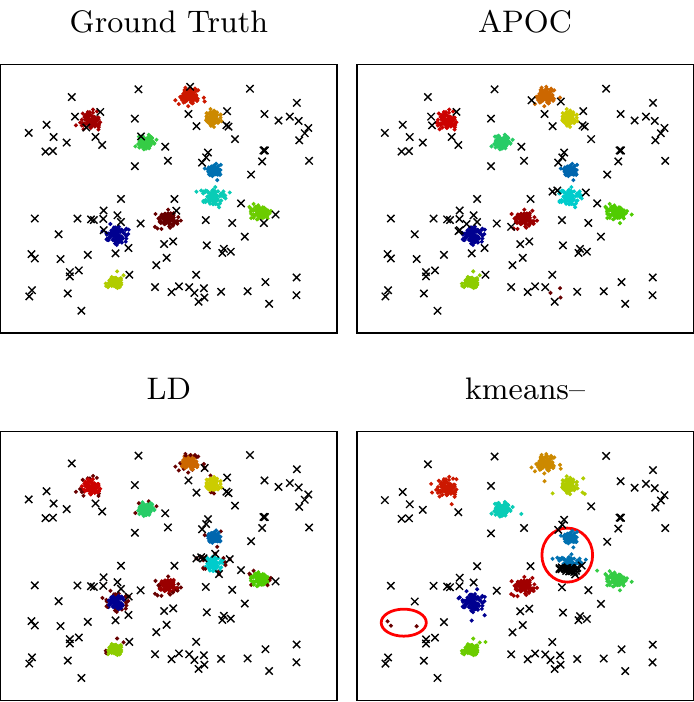}

    \caption{Clustering results for a 2D dataset with $k=10$, $100$
        points per cluster and $100$ outliers. APOC and LD accurately
        detect the clusters and select appropriate outliers without
        requiring $k$ as input. \kmmm{}, provided with the correct value
        of $k$, fails to split certain clusters which results in the
        selection of non-outliers as outliers. Black crosses indicate
        the outliers selected, points of identical colour indicate
        clusters and the red circles indicate errors made by \kmmm{}.}
    \label{fig:synth-2d-example}
    \vspace{-5mm}
\end{figure}

Clustering and outlier detection are often studied and investigated as
two separate problems \citep{chandola}. However, it is natural to
consider them simultaneously. For example, outliers can have a
disproportionate impact on the location and shape of clusters which in
turn can help identify, contextualise and interpret the outliers.

A branch of statistics known as ``robust statistics'' studies the design
of statistical methods which are less sensitive to the presence of
outliers \citep{huber_2008}. For example, the median and trimmed mean
estimators are far less sensitive to outliers than the mean. Similarly,
versions of Principal Component Analysis (PCA) have been proposed
\citep{croux_1996, wright_2009} which are more robust against model
mis-specification. An important primitive in the area of robust
statistics is the notion of Minimum Covariance Determinant (MCD): Given
a set of $n$ multivariate data points and a parameter $\ell$, the
objective is to identify a subset of points which minimises the
determinant of the variance-covariance matrix over all subsets of size
$n -\ell$. The resulting variance-covariance matrix can be integrated
into the Mahalanobis distance and used as part of a chi-square test to
identify multivariate outliers \citep{rousseeuw_1999}.

In the theoretical computer science literature, similar problems have
been studied in the context of clustering and facility location. For
example, \citet{chen_2008} has considered and proposed a constant factor
approximation algorithm for the k-median with outliers problem: Given
$n$ data points and parameters $k$ and $\ell$, the objective is to
remove a set of $\ell$ points such that the cost of k-median clustering
on the remaining $n - \ell$ points is minimised. Earlier
\citet{charikar_2001}, have proposed a bi-criteria approximation
algorithm for the facility location with outliers problem. While of
theoretical interest, none of these algorithms are amenable to a
practical implementation on large data sets.

Robustness of clustering methods in general is discussed in
\citep{garcia_2010, hennig_2008}. They provide a good theoretical
overview about the conditions under which clustering methods can deal
with noise or outliers. However, it is difficult to determine a-priori
if data exhibits the required properties.

More recently, \citet{chawla_2013} have proposed \klmeansalgo{} a
practical and scalable algorithm for the k-means with outlier problem.
\kmmm{} is a simple extension of the \kmeansalgo{} algorithm and is
guaranteed to converge to a local optima. However, the algorithm
inherits the weaknesses of the classical \kmeansalgo{} algorithm. These
are: (i) the requirement of setting the number of clusters $k$ and (ii)
initial specification of the $k$ centroids. It is well known that the
choice of $k$ and initial set of centroids can have a large impact on
the result. Trimmed \kmeansalgo{} \citep{cuesta_1997} is a special case
of \kmmm{} with $k=1$.

In this paper we present methods that approach the problem of joint
clustering and outlier detection as an integer programming optimisation
task. The resulting algorithms find the number of cluster on their own
and require as sole input the distance between pairs of points as well
as the number $\ell$ of outliers to select. We propose three methods to
solve this, each with their benefits and drawbacks: (i) affinity
propagation \citep{frey_2007} extension to outlier detection, (ii)
linear programming and (iii) Lagrangian duality relaxation. The main
contributions of the paper are as follows:
\begin{itemize}
    \item formulation of the clustering with outlier detection problem
        as an integer program;
    \item modification of the affinity propagation model and derivation
        of new update rules;
    \item scalable algorithm based on Lagrangian duality;
    \item approximation analysis of real data based on linear
        programming solutions;
    \item evaluation on synthetic and real-world data sets.
\end{itemize}

The remainder of the paper is structured as follows. In
\secref{problem-formulation} we describe the problem in detail.
Following that in \secref{methods} we describe the different methods. In
\secref{experiments} we evaluate the methods on both synthetic and real
datasets. \Secref{related-work} provides some additional information on
related work before we conclude in \secref{conclusion}.

\section{Problem Formulation}
\label{sec:problem-formulation}

Given the assignment cost matrix $d_{ij}$ and cluster creation costs $c_i$
we define the task of clustering and outlier selection as the problem of
finding the assignments to the binary exemplar indicators $y_i$, outlier
indicators $o_i$ and point assignments $x_{ij}$ that minimises the
following energy function:
\begin{equation}
    \min \sum_i c_i y_i + \sum_i \sum_x d_{ij} x_{ij} ,
    \label{eq:cost-function}
\end{equation}
subject to
\begin{align}
    x_{ij}              & \leq y_j \label{eq:ip-c-1} \\
    o_i + \sum_j x_{ij} & = 1      \label{eq:ip-c-2} \\
    \sum_i o_i          & = \ell   \label{eq:ip-c-3} \\
    x_{ij}, y_j, o_i    & \in \{0, 1\} \label{eq:ip-c-4}.
\end{align}
In order to obtain a valid solution a set of constraints have been
imposed:
\begin{itemize}
    \item points can only be assigned to valid exemplars \eqref{ip-c-1}
    \item every point must be assigned to exactly one other point or
        declared an outlier \eqref{ip-c-2}
    \item exactly $\ell$ outliers have to be selected \eqref{ip-c-3}
    \item only integer solutions are allowed \eqref{ip-c-4}
\end{itemize}
These constraints describe the facility location problem with outlier
selection (FLO). This allows the algorithm to select the number of clusters
automatically and implicitly defines outliers as those points whose
presence in the dataset has the biggest negative impact on the overall
solution.

In the following we explore three different methods of formulating and
solving the above generic problem. In the experiments we evaluate each
of the methods under different criteria such as quality of the solution,
complexity of the method and applicability to large datasets

\section{Methods}
\label{sec:methods}

In the following we will describe three ways of solving the problem
stated in \secref{problem-formulation}. We start with a linear
programming formulation which is known to provide the optimal answer if
the solution is integer. Next we propose an extension to affinity
propagation which is a conceptually nice method. Finally, we propose a
method based on Lagrangian duality which is highly scalable while
achieving results very close to the optimum found by LP.

\subsection{Linear Programming Relaxation}
\label{sec:lp}

The first method we present is based on linear programming and will
serve as ground truth for the other two methods. In order to solve the
integer program described in \secref{problem-formulation} we have to
relax it. If the solution to this relaxed formulation is integer, i.e.
all assignments are either $0$ or $1$, we have found the optimal
solution. The relaxed linear program has the following form:
\begin{equation}
    \min \sum_i \sum_j x_{ij} d_{ij} \label{eq:lp-energy}
\end{equation}
subject to
\begin{align}
    x_{ij}                       & \leq x_{ii} \label{eq:lp-c-1} \\
    \sum_j x_{ij}+\sum_k o_{ik}  & = 1         \label{eq:lp-c-2} \\
    \sum_i o_{ik}                & = 1         \label{eq:lp-c-3} \\
    0 \leq x_{ij}, o_{ik} \leq 1 & \hspace{5mm} \forall i, j, k
    \label{eq:lp-c-4}
\end{align}
where $d_{ij}$ is again the distance between point $i$ and $j$. This
follows \citet{komodakis_2008} with additional constraints to enforce
the outlier selection. We note here that \eqref{lp-energy} uses the
diagonal to indicate exemplar selection as it maps more easily to
affinity propagation. The constraint in \eqref{lp-c-1} enforces the
condition that points are only assigned to valid exemplars.
\eqref{lp-c-2} enforces that a point is either assigned to a single
cluster or is declared as an outlier. \eqref{lp-c-3} ensures that every
outlier point is used exactly once thus enforcing that exactly $\ell$
outliers are selected. Finally, \eqref{lp-c-4} is the relaxation of the
original integer program.

\subsection{Affinity Propagation Outlier Clustering}
\label{sec:apoc}

\begin{figure}[bt]
    \centering
    \includegraphics[width=\columnwidth]{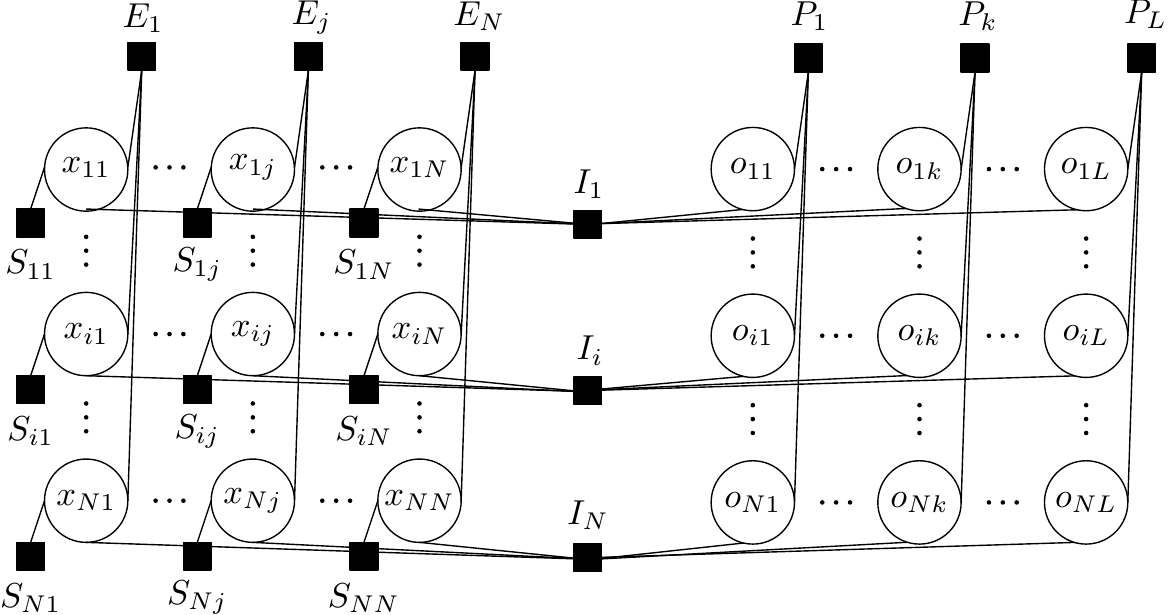}

    \caption{Graphical model of APOC. The left part is responsible for
    the clustering of the data, while the right part is responsible for
    the outlier selection. These two parts interact with each other via
    the $I$ factor nodes.}
    \label{fig:apoc-model}
\end{figure}

The extension to affinity propagation, based on the binary variable
model \citep{givoni_2009}, solves the integer program of
\secref{problem-formulation} by representing it as a factor graph, shown
in \figref{apoc-model}. This factor graph is solved using belief
propagation and is based on the following energy function:
\begin{multline}
    \max \sum_{ij} S_{ij}(x_{ij})
        + \sum_j E_j(x_{:j})
        + \sum_i I_i(x_{i:}, o_{i:})
        + \sum_k P_k(o_{:k}) ,
    \label{eq:apoc-objective-func}
\end{multline}
where
\begin{align}
    S_{ij}(x_{ij}) & = \begin{cases}
        -c_i    & \text{if } i = j \\
        -d_{ij} & \text{otherwise}
    \end{cases} \\
    I_{i}(x_{i:}, o_{i:}) & = \begin{cases}
        0       & \text{if } \sum_j x_{ij}+ \sum_k o_{ik}  = 1 \\
        -\infty & \text{otherwise}
    \end{cases} \\
    E_{j}(x_{:j}) & = \begin{cases}
        0       & \text{if } x_{jj} = \max_i c_{ij} \\
        -\infty & \text{otherwise}
    \end{cases} \\
    P_{k}(o_{:k}) & = \begin{cases}
        0       & \text{if } \sum_i o_{ik} = 1 \\
        -\infty & \text{otherwise}
    \end{cases}
\end{align}
with $x_{i:} = x_{i1}, \dots, x_{iN}$. Since we use the max-sum
algorithm we maximise the energy function and use negative distances.
The three constraints can be interpreted as follows:
\begin{enumerate}
    \item 1-of-$N$ Constraint ($I_i$). Each data point has to choose
        exactly one exemplar or be declared as an outlier.
    \item Exemplar Consistency Constraint ($E_j$). For point $i$
        to select point $j$ as its exemplar, point $j$ must declare
        itself an exemplar.
    \item Select $\ell$ Outliers Constraint ($P_k$). For every outlier
        selection exactly one point is assigned.
\end{enumerate}
These constraints are enforced by associating an infinite cost with
invalid configurations,thus resulting in an obviously suboptimal
solution.

\begin{figure}[bt]
    \centering

    \includegraphics[height=2.0cm]{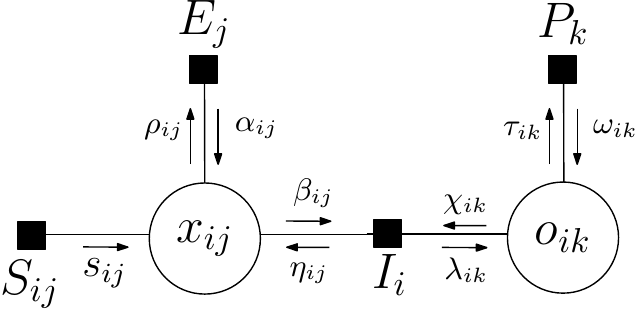}

    \caption{Messages exchanged by the APOC graphical model. $x_{ij}$
    represents the clustering choice whereas $o_{ik}$ represents the
    outlier choice.}
    \label{fig:apoc-messages}
    \vspace{-5mm}
\end{figure}

The energy function is optimised with the max-sum algorithm
\citep{kschischang_2001}, which allows the recovery of the maximum a
posteriori (MAP) assignments of the $x_{ij}$ and $o_{ik}$ variables. The
algorithm works by exchanging messages between nodes in the factor
graph. In their most general form these messages are defined as follows:
\begin{align}
    \label{eq:mp-nf}
    \mu_{v \rightarrow f}(x_v) & =
        \sum_{f^* \in ne(v) \setminus f} \mu_{f^* \rightarrow v}(x_v), \\
    \begin{split}
    \label{eq:mp-fn}
    \mu_{f \rightarrow v}(x_v) & =
        \max_{x_1, \dots, x_M} \Big[ f(x_v, x_1, \dots, x_M) \\
        & + \sum_{v^* \in ne(f) \setminus v} \mu_{v^* \rightarrow f}(x_{v^*}) \Big],
    \end{split}
\end{align}
where $\mu_{v \rightarrow f}(x)$ is the message sent from node $v$ to
factor $f$, $\mu_{f \rightarrow v}(x_v)$ is the message from factor $f$
sent to node $v$, $ne()$ is the set of neighbours of the given factor or
node and $x_v$ is the value of node $v$.

The messages exchanged by APOC are shown in \figref{apoc-messages}. We
can see that each node $x_{ij}$ is connected to three factors: $S_{ij}$,
$I_i$ and $E_j$ whereas outlier nodes $o_{ik}$ are connected to only
two, $I_i$ and $P_k$. Messages $\rho_{ij}$, $\beta_{ij}$, $\tau_{ik}$
and $\xi_{ik}$ are sent from nodes to factors and derived using
\eqref{mp-nf}. The other five messages $s_{ij}, \alpha_{ij}$,
$\eta_{ij}$, $\lambda_{ik}$ and $\omega_{ik}$ are derived with
\eqref{mp-fn} since they are sent from a factor to a node. Since only
binary variables are involved it is sufficient to compute the difference
between the two variable settings. Combining these messages we obtain
the final set of update equations as:

\begin{align}
    \rho_{ij}    & = s_{ij} + \min \left[
        -\max_{t \neq j}(\alpha_{it} + s_{it}),
            -\max_{t}(\omega_{it})
        \right]
    \label{eq:apoc_rho} \\
    \alpha_{ij}  & =
        \begin{cases}
            \sum_{t \neq j} \max(0, \rho_{tj}) & i = j \\
            \min \left[ 0, \rho_{jj} + \sum_{t \notin \{i, j\}} \max(0, \rho_{tj}) \right] & i \neq j
        \end{cases}
    \label{eq:apoc_alpha} \\
    \lambda_{ik} & = \min \left[
            -\max_t(\alpha_{it} + s_{it}),
            -\max_{t \neq k}(\omega_{ti})
        \right]
    \label{eq:apoc_lambda} \\
    \omega_{ik}  & = -\max_{t \neq i}(\lambda_{tk})
    \label{eq:apoc_omega}
\end{align}

The above equations show how to update the messages, however, we still
need to explain how to initialise the messages, determine convergence
and extract the MAP solution. First, it is important to set the diagonal
entries of $S$ properly. Typically using $S_{ii} = \theta * median(S)$
is a good choice, with $\theta \in [1, 30]$. The messages $\alpha_{ij},
\rho_{ij}$ and $\lambda_{ik}$ are initialised to $0$ and $\omega_{ik}$
to the median of $S$. Once the messages are initialised we update them
in turn with damping until we achieve convergence.  Convergence is
achieved when the energy of the solution is not changing drastically
over a few iterations. The outliers are determined as the $\ell$ points
with the largest values of $\max_k (\lambda_{ik} + \omega_{ik})$. From
the remaining points the exemplars are then selected as the points for
which $(\alpha_{ii} + \rho_{ii}) > 0$ is true. All other points $i$ are
assigned to the exemplar $e$ satisfying $\argmax_e(\alpha_{ie} +
\rho_{ie})$. This entire process is shown in \algoref{apoc-algo}, where
we first initialise the messages, then update them until convergence and
finally extract the MAP solution.

\begin{algorithm}[bt]
    \caption{$\text{apoc}(S, \ell)$}
    \label{algo:apoc-algo}

    \ForEach{$i,j \in \{1,\dots,N$\}}
    {
        $\alpha_{ij} \leftarrow 0$\;
        $\rho_{ij} \leftarrow 0$\;
    }
    \ForEach{$i \in \{1,\dots,N\}, k \in \{1,\dots,\ell\}$}
    {
        $\lambda_{ik} \leftarrow 0$\;
        $\omega_{ik} \leftarrow \text{median}(S)$\;
    }

    \Repeat{convergence}
    {
        update $\rho$ according to \eqref{apoc_rho}\;
        update $\alpha$ according to \eqref{apoc_alpha}\;
        update $\lambda$ according to \eqref{apoc_lambda}\;
        update $\omega$ according to \eqref{apoc_omega}\;
    }

    $O \leftarrow$ extract outliers\;
    $E \leftarrow$ extract exemplars\;
    $A \leftarrow$ find exemplar assignments\;
\end{algorithm}

\subsection{Lagrangian Duality}
\label{sec:lagrange}

The final method is based on Lagrangian duality. The basic idea is to
relax the original problem by introducing Lagrange multipliers
$\lambda$. The result is a dual problem which is now concave with a
unique maximum, making it possible to use gradient ascent based methods.
However, the resulting function is not differentiable and requires the
use of subgradients.

We restate the original problem here for convenience:
\begin{equation}
    \min \sum_{j} y_j c_j + \sum_{i}\sum_{j} x_{ij} d_{ij}
\end{equation}
subject to
\begin{align}
    x_{ij}              & \leq y_j \label{eq:ld-c-1} \\
    o_i + \sum_j x_{ij} & = 1      \label{eq:ld-c-2} \\
    \sum_i o_i          & = \ell   \label{eq:ld-c-3} \\
    0 \leq x_{ij}, y_j, o_i \leq 1 & \hspace{5mm} \forall i,j ,
\end{align}
where $o_i = 1$ indicates that point $i$ is an outlier and $\ell$ is the
number of outliers we wish to find. \eqref{ld-c-1} encodes that a point
can only be assigned to a valid exemplar. The second constraint
\eqref{ld-c-2} enforces that a point is either assigned to a single
cluster or selected as an outlier. The final constraint, \eqref{ld-c-3},
ensures that exactly $\ell$ outliers are selected. Relaxing
\eqref{ld-c-2} yields:
\begin{equation}
    \min
        \underbrace{\sum_i (1 - o_i) \lambda_i}_{\text{outliers}} +
        \underbrace{\sum_j c_j y_j + \sum_i \sum_j (d_{ij} - \lambda_i)
        x_{ij}}_{\text{clustering}} .
    \label{eq:ld-relaxed}
\end{equation}
subject to
\begin{align}
    x_{ij}      & \leq y_i \\
    \sum_k o_k  & = \ell   \\
    0 \leq x_{ij}, y_j, o_k \leq 1 & \hspace{5mm} \forall i,j,k
\end{align}
We now solve this relaxed problem using the idea of
\citet{bertsimas_2005} by finding valid assignments that attempt to
minimise \eqref{ld-relaxed} without optimality guarantees. The Lagrange
multipliers $\lambda$ act as a penalty incurred for constraint violation
which we try to minimise. From \eqref{ld-relaxed} we see that the
penalty influences two parts: outlier selection and clustering.  We
select good outliers by designating the $\ell$ points with largest
$\lambda$ as outliers, as this removes a large part of the penalty. For
the remaining $N - \ell$ points we determine clustering assignments by
setting $x_{ij} = 0$ for all pairs for which $d_{ij} - \lambda_i \geq
0$. To select the exemplars we compute
\begin{equation}
    \mu_j = c_j + \sum_{i : d_{ij} - \lambda_i < 0} (d_{ij} -
    \lambda_i),
\end{equation}
which represents the amortised cost of selecting point $j$ as exemplar
and assigning points to it. Thus, if $\mu_j < 0$ we select point $j$ as
an exemplar and set $y_j = 1$, otherwise we set $y_j = 0$. Finally, we
set $x_{ij} = y_j$ if $d_{ij} - \lambda_i < 0$. From this complete
assignment we then compute a new subgradient $\vec s^t$ and update the
Lagrangian multipliers $\lambda^t$ as follows:
\begin{align}
    \vec s_j^t  & = 1 - \sum_j x_{ij} \\
    \lambda_j^t & = \max(\lambda_j^{t-1} + \theta^t \vec s_j, 0) ,
\end{align}
where $\theta^t$ is the step size at time $t$ computed as
\begin{equation}
    \theta^t = \theta^0 \: \text{pow}(\alpha, t) \quad \alpha \in (0, 1) ,
\end{equation}
where $\text{pow}(a, b) = a^b$.  To obtain the final solution we repeat
the above steps until the changes become small enough, at which point we
extract a feasible solution. This is guaranteed to converge
\citep{bertsimas_2005} if a step function is used for which the
following holds:
\begin{equation}
    \sum_{t=1}^{\infty} \phi_t = \infty
    \qquad \text{and} \qquad
    \lim_{t\rightarrow\infty} \theta_t = 0 .
\end{equation}

\subsubsection{Scalable Implementation Considerations}

In order to enable the algorithm to scale to large datasets we have to
consider the limited availability of main memory. First we cannot assume
that the complete distance matrix can fit into main memory.  Therefore,
we compute the distances on the fly. Since this involves $N^2$
evaluations per iteration it is the most costly part of the method.
However, the evaluation of the distance function can be easily
parallelised. In practice with simple distance functions, such as the
Euclidean distance, approximately 75\% of the computational time is
spent evaluating the distance function. Another important point is that
just as storing the full distance matrix is not possible, neither is
storing the full assignment matrix $\vec x$. However, we are only
interested in the values where $x_{ij} = 1$, which is a small portion of
the full matrix. Thus we can use standard sparse matrix implementations
to manage the assignment matrix. 

The pseudo code in \algoref{ld-algo} shows how to compute the assignment
matrix $\vec x$ using the above mentioned improvements. Lines 1 through
5 initialise the required storage. In lines 6 to 12 we sort the points
in descending order according to their $\lambda$ values, by first
creating pairs of point index and value and then sorting these. The
computationally intensive but also parallelisable part of the algorithm
is located in lines 13 to 21. There we compute the exemplar score
$\mu_j$ and at the same time remember point pairs $(i, j)$ for which
$d_{ij} - \lambda_i < 0$. Lines 22 to 27 then perform the exemplar
assignments based on $\mu_j$. Finally, in lines 28 to 31 we set the
assignment $x_{ij} = 1$ if $y_j = 1$. Lastly, we return assignments,
exemplars and outliers.

\begin{algorithm}[bt]
    \SetAlgoLined

    \caption{$\text{LD-Iteration}(\lambda)$}
    \label{algo:ld-algo}

    $\mathcal O \leftarrow \vec 0$\tcp*{Outlier indicators}
    $\vec y \leftarrow \vec 0$\tcp*{Exemplar indicators}
    $\mathcal L \leftarrow \emptyset$\tcp*{Set of $(i, \lambda_i)$ pairs}
    $\mathcal S \leftarrow \emptyset$\tcp*{Assignment pairs $(i, j)$}
    $\vec x \leftarrow \vec 0$\tcp*{Assignments}

    \tcp{Selecting outliers}
    \ForEach{$i \in \{1,\dots,N\}$}
    {
        $\mathcal L \leftarrow \mathcal L \cup (i, \lambda_i)$;
    }
    $\mathcal L \leftarrow sort(\mathcal L)$\;
    \ForEach{$i \in \{1,\dots,L\}$}
    {
        $\vec \mathcal O_{\mathcal L_{i,1}} \leftarrow 1$\;
    }

    \tcp{Compute exemplar and outlier scores}
    \ForEach{$j \in \{1,\dots,N$\}}
    {
        $\mu_j  \leftarrow c_j$\;
        \ForEach{$i \in \{1,\dots,N$\}}
        {
            $q \leftarrow dist(i, j) - \lambda_i$\;
            \uIf{$q < 0$}
            {
                $\mu_j \leftarrow \mu_j + q$\;
                $\mathcal S \leftarrow \mathcal S \cup (i, j)$\;
            }
        }
    }

    \tcp{Select exemplars}
    \ForEach{$j \in \{1,\dots,N$\}}
    {
        \eIf{$\mu_j < 0$}
        {
            $\vec y_j \leftarrow 1$\;
        }
        {
            $\vec y_j \leftarrow 0$\;
        }
    }

    \tcp{Perform cluster assignments}
    \ForEach{$s \in \mathcal S$}
    {
        \If{$\mu_{s_{2}} = 1$}
        {
            $x_{p_{1}, p_{2}} \leftarrow 1$\;
        }
    }

    \Return{$\vec x, \vec y, \mathcal O$}\;
\end{algorithm}

\subsection{Comparisons}
\label{sec:method-comparison}

Now that we have presented the different methods we want to give an
overview of the advantages and disadvantages of these.
\tabref{method-comparison} presents a quick overview of different
properties of the proposed methods discussed in more detail next. First,
the speed of APOC and LD clearly outperform LP by a large margin,
however, other methods such as \kmmm{} still perform better.

With regards to optimality we know that
\begin{equation}
    FLO_{LP} \penalty 0 \leq FLO_{LD} \leq FLO_{IP}
\end{equation}
\citep{bertsimas_2005}. Additionally, if LP finds a solution that is
integer it is equivalent to the IP one, i.e. $FLO_{LP} = FLO_{IP}$. In
the experiments we see that even though LP finds the optimal solution, LD
fails to do so. This contradicts theory, but is explained by the fact
that for performance gains we sacrifice some optimality. More
specifically, we use discounted updates of the solution matrix $\vec x$
as well as stop before we converge to a pure integer solution. APOC has
no theoretical guarantees on either convergence or optimality bounds of
the solution. While such guarantees can be given for certain types of
structures with belief propagation \citep{weiss_2007} it is unclear how
they apply to the special case of APOC. However, in practice both APOC
and LD achieve scores very close to the optimum.

The speed of LP clearly makes it impossible to scale for large datasets
which leaves us with the comparison of APOC and LD. APOC requires
storage for messages and similarities and operations which cannot be
supported by standard sparse matrix implementations. As such APOC cannot
be made truly scalable. LD on the other hand spends the majority of its
runtime computing distances between pairs of points, a task that can
easily be parallelised. Furthermore the actual assignment matrix can
easily be stored using standard sparse matrix implementations. Thus LD
is a much better candidate for large scale datasets. Finally, with
regards to extensibility LP is the easiest as the relaxation only
touches the variables. LD is more involved as there is more freedom in
the relaxation of the constraints and a way to solve the dual problem
efficiently has to be devised. Affinity propagation is the hardest one
as it requires very careful choice of constraints which can be modelled
by the graphical model and laborious derivation of update rules, which
makes it much harder to come up with solutions to new problems.

\begin{table}[bt]
    \caption{Advantages and disadvantages of LP, APOC and LD. See
        \secref{method-comparison} for detailed explanation.}
    \label{tab:method-comparison}

    \centering
    \normalsize

    \begin{tabular}{lccc}
        \toprule
                        & LP  & APOC & LD  \\
        \midrule
        Speed           & $--$ & $+$  & $+$ \\
        Optimality      & $++$ & $+$  & $+$ \\
        Scalability     & $--$ & $-$  & $+$ \\
        Extensibility   & $++$ & $-$  & $+$ \\
        \bottomrule
    \end{tabular}
    \vspace{-5mm}
\end{table}

\section{Experiments}
\label{sec:experiments}

In this section the proposed methods are evaluated on both synthetic and
real data. We first present experiments using synthetic data to show
different properties of the presented methods and a quantitative
analysis. We then process GPS traces of hurricanes to show the
applicability of this method to real data. Finally, we present results
on two image datasets visually showing the quality of the clusters and
outliers found.

Both APOC and LD require a cost for creating clusters. In all
experiments we obtain this value as $\theta * \text{median}(x_{ij})$,
i.e. the median of all distances multiplied by a scaling factor $\theta$
which is typically in the range $[1, 30]$.

\subsection{Synthetic Data}

We use synthetic datasets to evaluate the performance of the
proposed methods in a controlled setting. We randomly sample $k$ clusters
with $m$ points each from $d$-dimensional normal distributions $\mathcal
N(\mu, \Sigma)$ with randomly selected $\mu$ and $\Sigma$. To these
clusters we add $\ell$ additional outlier points that have a low
probability of belonging to any of the selected clusters.

We compare APOC and LD against \kmmm{} using k-means{+}{+}
\citep{arthur_2007} to select the initial centres. Euclidean distance is
used as the metric for all methods. Unless mentioned otherwise \kmmm{}
is provided with the exact number of clusters generated, while APOC and
LD are required to determine this automatically.

To assess the performance of the methods we use the following three
metrics:
\begin{enumerate}
    \item Normalised Jaccard index, measures how accurately a method
        selects the ground truth outliers. It is a coefficient computed
        between selected outliers $O$ and  ground-truth outliers $O^*$.
        The final coefficient is normalised with regards to the best
        possible coefficient obtainable in the following way:
        \begin{equation}
            J(O, O^*) = \frac{|O \cap O^*|}{|O \cup O^*|} /
            \frac{\min(|O|, |O^*|)}{\max(|O|, |O^*|)} .
        \end{equation}

    \item Local outlier factor \citep{breunig_2000} (LOF) measures the
        outlier quality of a point. We compute the ratio between the
        average LOF of $O$ and $O^*$, which indicates the quality of the
        set of selected outliers.

    \item V-Measure \citep{rosenberg_2007} indicates the quality of the overall
        clustering solution. The outliers are considered as an
        additional class for this measure.
\end{enumerate}
For all the metrics a value of 1 is the optimal outcome.

In \figref{synth-2d-example} we present clustering results obtained by
the different methods. Both APOC and LD manage to identify the correct
number of clusters and select accurate outliers. \kmmm{} on the other
hand, even with good initialisation and correct value for $k$ specified,
fails to find the correct clusters and as a result finds a suboptimal
solution. The errors made by \kmmm{} are highlighted by the red circles.

\begin{figure*}[bt]
    \centering

    \includegraphics{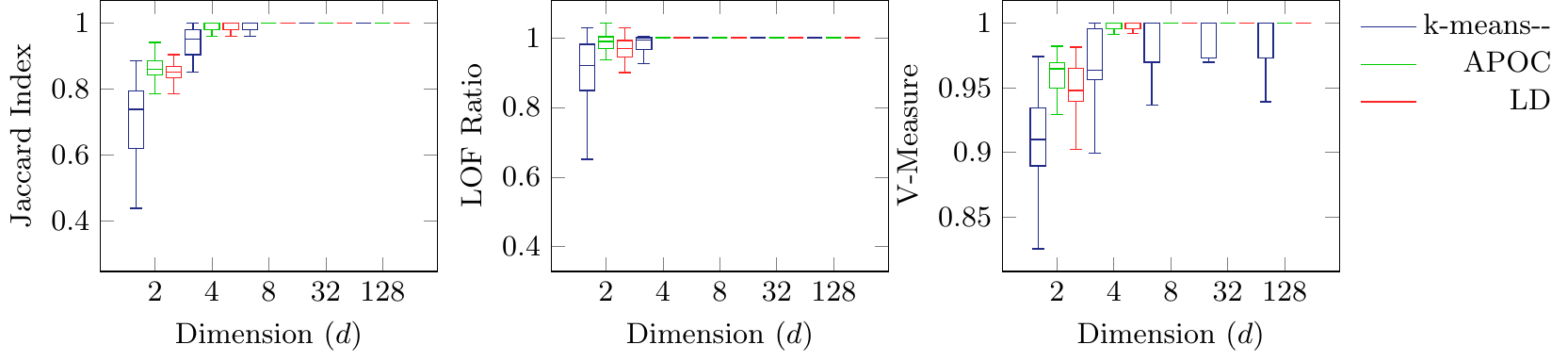}

    \caption{The impact of increasing the data dimensionality on the
        quality of the clustering and outlier selection quality. APOC
        and LD provide similar results while \kmmm{}, provided with the
        correct $k$, has more trouble.}
    \label{fig:synth-dimensions}
\end{figure*}

We first investigate the influence of the data dimensionality on the
results. From \figref{synth-dimensions} it is clear that in general the
quality of the solution increases with higher dimensions. This can be
explained by the fact that in higher dimensional spaces the points are
farther apart and hence easier to cluster. Looking at \kmmm{} we can see
that it struggles more then the other two methods even though it is
provided with the correct number of clusters. In higher dimensions it
achieves perfect scores for the two outlier centric measures but is
unable to always find the correct solution to the whole clustering
problem. Looking at APOC and LD we can see that both have little trouble
finding perfect solutions in high dimensions. In lower dimensions LD
shows a bit more variability compared to APOC. In general, the two
dimensional data is the most challenging one and thus will be used for
all further experiments.

\begin{figure*}[bt]
    \centering

    \includegraphics{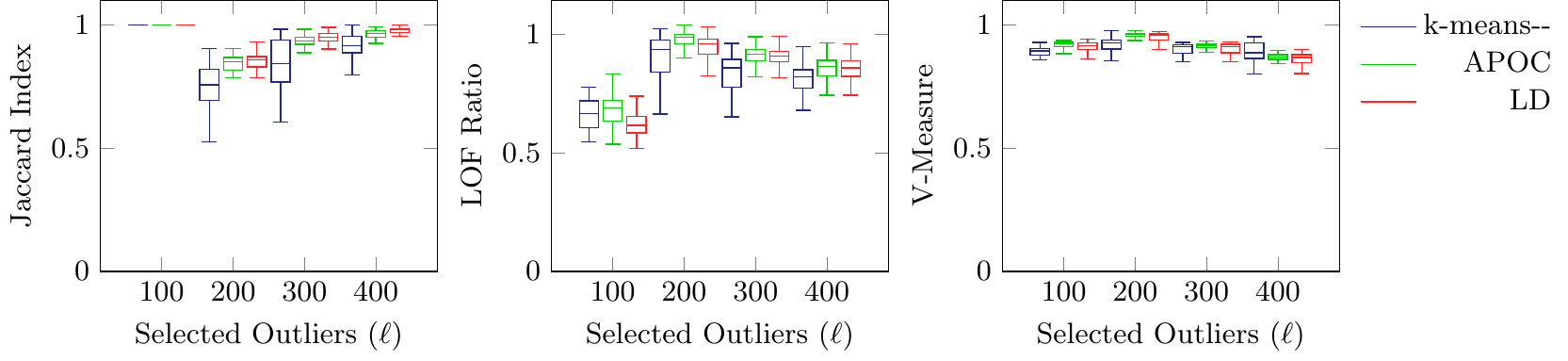}

    \caption{The impact of number of outliers specified ($\ell$) on the
        quality of the clustering and outlier detection performance.
        APOC and LD perform similarly with more stability and better
        outlier choices compared to \kmmm{}. We can see that
        overestimating $\ell$ is more detrimental to the overall
        performance, as indicated by the LOF Ratio and V-Measure,
        then underestimating it.}
    \label{fig:synth-outliers}
\end{figure*}

The number of outliers $\ell$ is a parameter that all methods require.
Typically the correct value of $\ell$ is unknown and it is therefore
important to know how the algorithms react when the user's estimate is
incorrect. We generate 2D datasets with 2000 inliers and 200 outliers
and vary the number of outliers $\ell$ selected by the methods.  The
results in \figref{synth-outliers} show that in general no method breaks
completely if the correct value for $\ell$ is not provided.  Looking at
the Jaccard index we see that if $\ell$ is smaller then the true number
of outliers all methods pick only outliers. When $\ell$ is greater then
the true number of outliers we can see a difference in performance,
namely that LD picks the largest outliers which APOC does to some extent
as well, while \kmmm{} does not seem to be very specific about which
points to select. This difference in behaviours stems from the fact that
APOC and LD optimise a cost function in which removing the biggest
outliers is the most beneficial. The LOF ratio shows a similar picture
as the Jaccard index. For $\ell = 100$ the values are much lower as not
all outliers can be picked. For the other cases APOC and LD perform
similarly while \kmmm{} shows higher variability. Finally, V-Measure
shows that the overall clustering results remain accurate even if the number
of outliers is misspecified. For large differences between actual and
specified outliers a drop in clustering performance can be observed.

\begin{figure*}[bt]
    \centering

    \subfloat[Speedup over LP]{\includegraphics{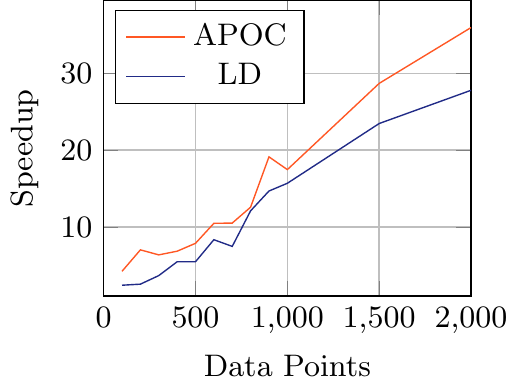}}
    \subfloat[Total Runtime]{\includegraphics{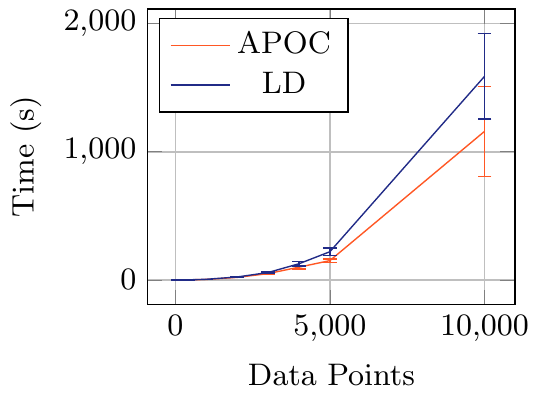}}
    \subfloat[Time per Iteration]{\includegraphics{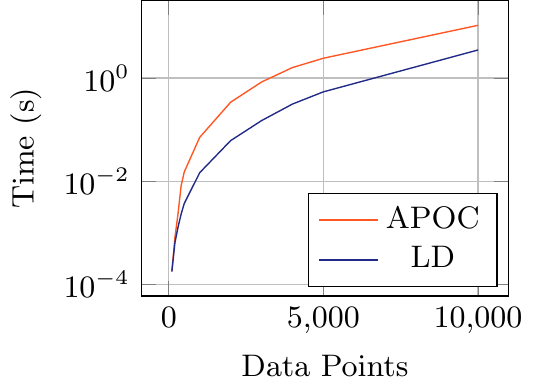}}

    \caption{The graphs shows how the number of points influences
        different measures. In (a) we compare the speedup of both APOC
        and LD over LP. (b) compares the total runtime needed to solve
        the clustering problem for APOC and LD. Finally, (c) plots the
        time required (on a log scale) for a single iteration for APOC and
        LD.}

    \label{fig:lp-ld-apoc-comparisons}
    \vspace{-5mm}
\end{figure*}

Since both APOC and LD are not guaranteed to find the optimal solution
we investigate how close the solutions obtained are to the optimum. The
ground truth needed for this is obtained by solving the LP formulation
of \secref{lp} with CPLEX. This comparison indicates what quality 
can be typically expected from the two methods. Additionally, we can evaluate
the speed of these approximations. We evaluate 100 datasets,
consisting of 2D Gaussian clusters and outliers, with varying number of
points. On average APOC obtains an energy that is $96\% \pm 4\%$
of the optimal solution found by LP, LD obtains $94\% \pm 5\%$ of the LP
energy while \kmmm{}, with correct k, only obtains $86\% \pm 12\%$ of
the optimum. These results reinforce the previous analysis;
APOC and LD perform similarly while outperforming \kmmm{}. We now
look at the speedup of APOC and LD over LP, as shown in
\figref{lp-ld-apoc-comparisons} a). Both methods outperform LP by a
large margin which only grows the more points are involved. Overall for
a small price in quality the two methods obtain a solution significantly
faster.

Next we compare the performance between APOC and LD.
\figref{lp-ld-apoc-comparisons} b) shows the overall runtime of both
methods for varying number of data points. Here we observe that APOC
takes less time then LD. However, by observing
the time a single iteration takes, shown in
\figref{lp-ld-apoc-comparisons} c), we see that LD is much faster on a
per iteration basis compared to APOC. In practice LD requires several
times the number of iterations of APOC, which is affected by
the step size function used. Using a more sophisticated
method of computing the step size will provide large gains to LD.
Finally, the biggest difference between APOC and LD is that the former
requires all messages and distances to be held in memory. This
obviously scales poorly to large datasets. Conversely, LD computes the
distances during running time and only needs to store indicator vectors and a
sparse assignment matrix, thus using much less memory. This makes LD
amenable to processing large scale datasets. For example, with
single precision floating point numbers, dense matrices and \num{10000}
points APOC requires around \SI{2200}{\mega\byte} of memory while LD
only needs \SI{370}{\mega\byte}. Further gains can be obtained by 
using sparse matrices which is straight forward in the case of LD but complicated for
APOC.

\subsection{Hurricane Data}

\begin{figure*}[bt]
    \centering
    \subfloat[Outliers]{\includegraphics[width=0.3\linewidth]{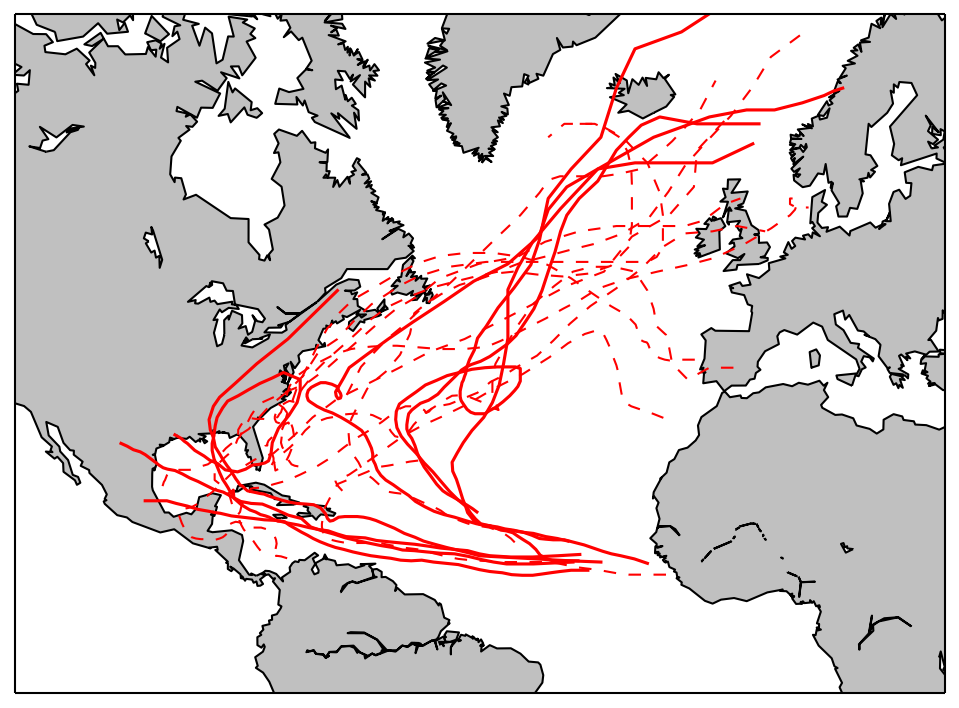}}
    \subfloat[Clusters]{\includegraphics[width=0.3\linewidth]{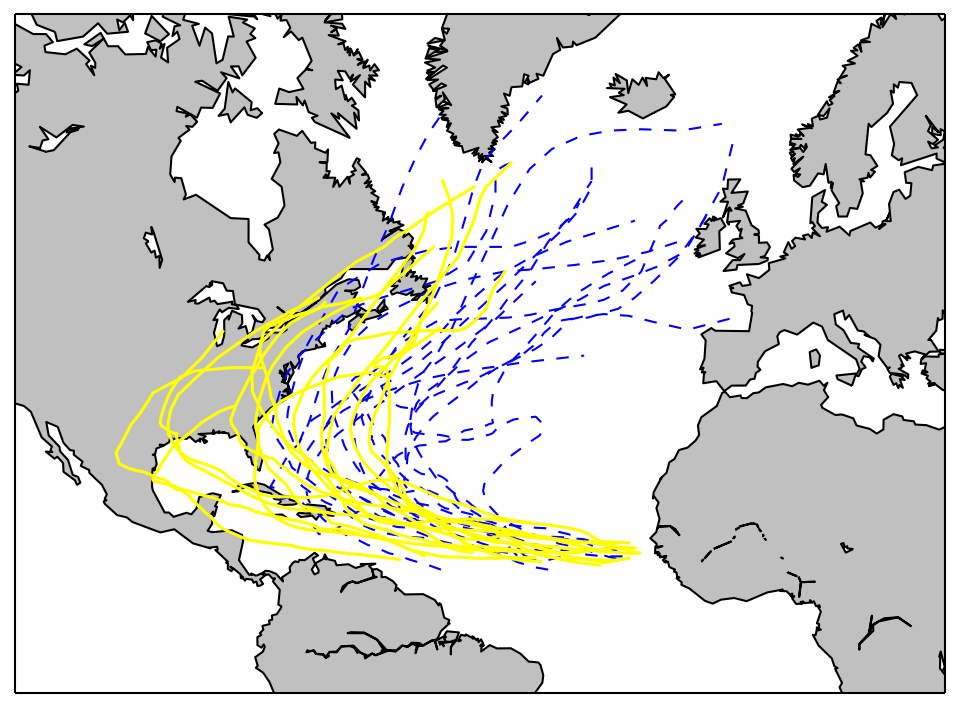}}
    \subfloat[Clusters]{\includegraphics[width=0.3\linewidth]{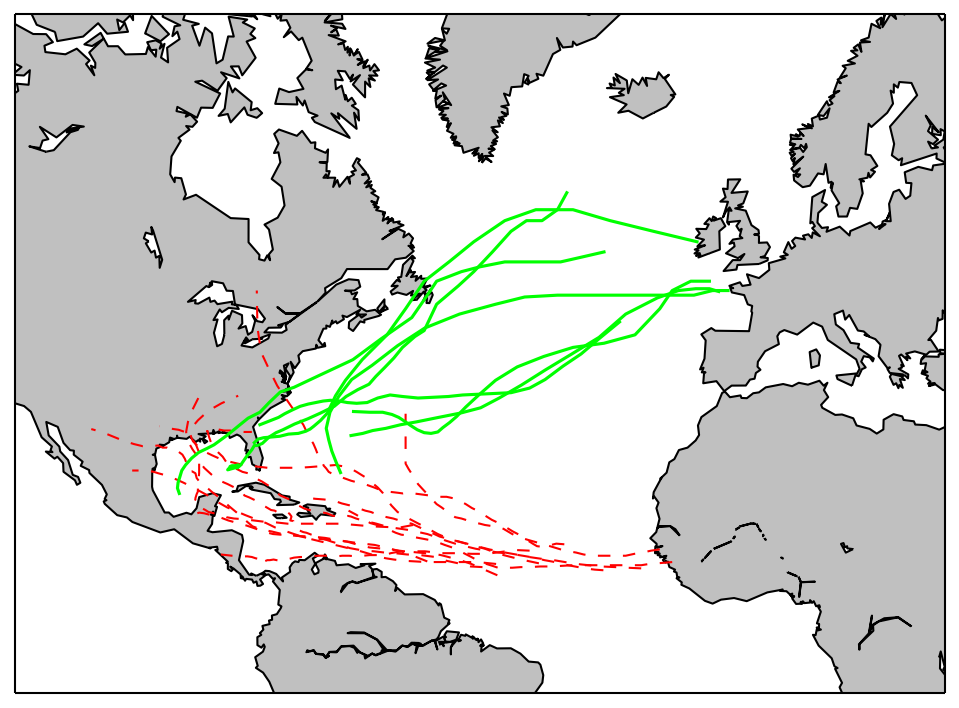}}

    \caption{Outliers and clusters found in the hurricane data set.
        In (a) we show the outliers as two groups: Cape Verde-type
        hurricanes (thick lines) and others (dashed lines). In (b) and
        (c) we display examples of clusters which all exhibit similar
        shape albeit not necessarily in the same position.}
    \label{fig:hurricanes_1970}
\end{figure*}

We use hurricane data from 1970 to 2010 provided by the National Oceanic
and Atmospheric Administration (NOAA) in this experiment. The data
provides a time series of longitude and latitude coordinates for each
storm, effectively forming a GPS trace. In order to compare the overall
shape of these traces we use the discrete Fr\'echet distance
\citep{eiter_1994}.  Intuitively this measures the minimal distance
required to connect points on two curves being compared, i.e.
structurally similar curves have a low score. Before computing the
Fr\'echet distance between two curves we move their starting points to
the same location. This means that we are comparing their shapes and not
their global location. Clustering the \num{700} traces using the LD
method with $\ell = 20$ we obtain clusters that move in a similar
direction, as shown in \figref{hurricanes_1970} b) and c). Analysing the
outliers shown in \figref{hurricanes_1970} a) we find that half of them
are category 4 and 5 Cape Verde-type hurricanes, shown by the thick
stroke. The other half of the outliers were selected due to either their
long life time, unusual motion, or high destructive power and are shown
with dashed stroke. This demonstrates that the outliers found by the
proposed methods find interesting patterns in spatial temporal data
which are not directly apparent.

\begin{figure}[bt]
    \centering

    \subfloat[Exemplars]{
        \includegraphics{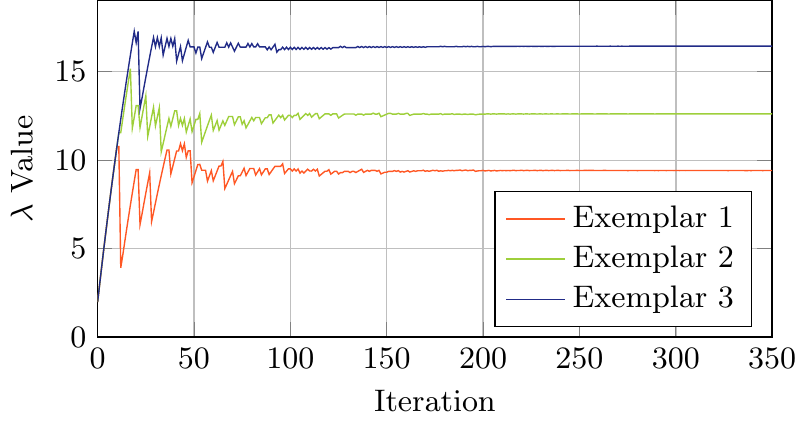}
        \label{fig:hurricane-evolution-outliers}
    } \\
    \subfloat[Outliers]{
        \includegraphics{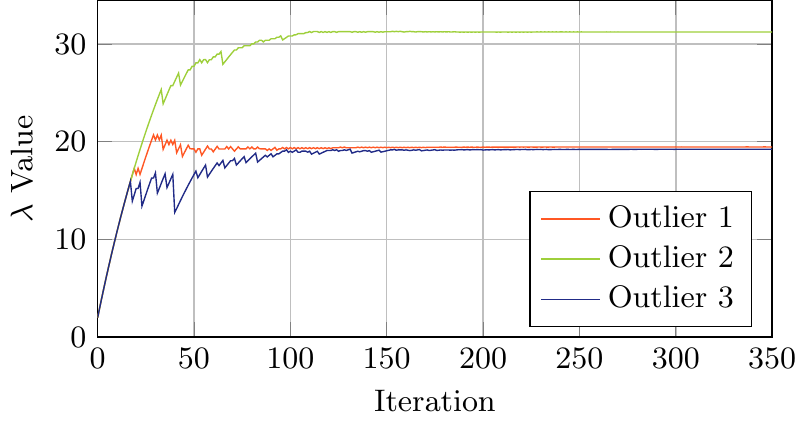}
        \label{fig:hurricane-evolution-lambda}
    }

    \caption{Evolution of the $\lambda$ values of select exemplars and
        outliers over the iterations. In both cases we can observe how
        after about 150 iterations the values have stabilised.
        Additionally, outliers have higher $\lambda$ values and exhibit
        less variation compared to exemplars.}
    \label{fig:ld-evolution}
\end{figure}

To better understand the behaviour of LD we plot the value of the
$\lambda$ values associated with representative outliers and exemplars
in \figref{ld-evolution}. One can see how after about 100 iterations the
values level out and remain stable. Interestingly the outliers have
higher $\lambda$ values compared to the exemplars which allows the
method to distinguish between them more easily. Lastly, the outliers
have smoother curves, i.e. less jumps, when compared to the exemplars.
This indicates that outliers do not change significantly whereas
exemplars will change during the optimisation process. Overall, stable
results are achieved after about a third of the runtime suggesting that
a more sophisticated termination criterion can be used for faster
convergence.

\subsection{MNIST Data}

\begin{figure}[bt]
    \centering

    \subfloat[Digit 1]{\includegraphics[width=0.32\columnwidth]{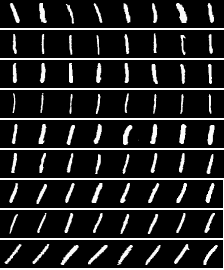}}
    \hfill
    \subfloat[Digit 4]{\includegraphics[width=0.32\columnwidth]{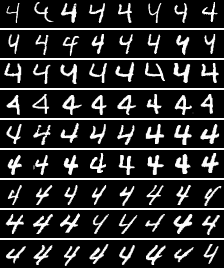}}
    \hfill
    \subfloat[Outliers]{\includegraphics[width=0.32\columnwidth]{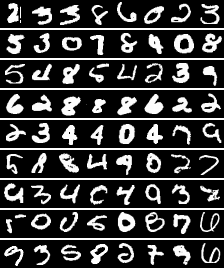}}

    \caption{Each row in (a) and (b) shows a different appearance of a
    digit captured by a cluster. The outliers shown in (c) tend to have
    heavier then usual stroke, are incomplete or are not recognisable as a
    digit.}
    \label{fig:mnist}
\end{figure}

The MNIST dataset, introduced by \citet{lecun_1998}, contains $28 \times
28$ pixel images of handwritten digits. We extract features from these
images by representing them as 768 dimensional vectors which is reduced
to 25 dimensions using PCA. $L2$ norm is used to compute the distance
between these vectors. In \figref{mnist} we show exemplary results
obtained when processing \num{10000} digits with the LD method with
$\theta = 5$ and $\ell = 500$. Each row in \figref{mnist} a) and b)
shows examples of clusters representing the digits 1 and 4 respectively.
This illustrates how different the same digit can appear and the
separation induced by the clusters.  \Figref{mnist} c) contains a subset
of the outliers selected by the method. These outliers have different
characteristics that make them sensible outliers, such as: thick stroke,
incomplete, unrecognisable or ambiguous meaning.

To investigate the influence the cluster creation cost has we run the
experiment with different values of $\theta$. In \tabref{mnist-evaluation}
we show results for values of cost scaling factor $\theta = \{5, 15,
25\}$. The V-Measure score does not change drastically between the
different runs.  However, the other measures, especially the number of
clusters changes.  We can see that, as expected, by increasing the cost
the number of clusters decreases. This results in an increased
completeness score, i.e. a larger percentage of the same ground truth
labels are contained in a smaller number of clusters. Meanwhile, the
homogeneity score drops which indicates that there are more cases of
different digits that look similar, such as $1$ and $7$, being placed
into a single cluster. Overall, we can say that changes to the cluster
creation cost has a predictable impact on the results with no apparent
sudden changes.

\begin{table}[bt]
    \caption{Evaluation of clustering results of the MNIST data set with
        different cost scaling values $\theta$. We can see that increasing
        the cost results in fewer clusters but as a trade off reduces
        the homogeneity of the clusters.}
    \label{tab:mnist-evaluation}

    \centering
    \normalsize

    \begin{tabular}{lrrr}
        \toprule
        $\theta$        & $5$    & $15$   & $25$  \\
        \midrule
        V-Measure       & $0.51$ & $0.53$ & $0.53$ \\
        Homogeneity     & $0.82$ & $0.73$ & $0.70$ \\
        Completeness    & $0.37$ & $0.42$ & $0.43$ \\
        Clusters        & $82$   & $36$   & $21$ \\
        \bottomrule
    \end{tabular}
    \vspace{-5mm}
\end{table}

\subsection{Natural Scenes Data}

\begin{figure}[bt]
    \centering

    \includegraphics[width=\columnwidth]{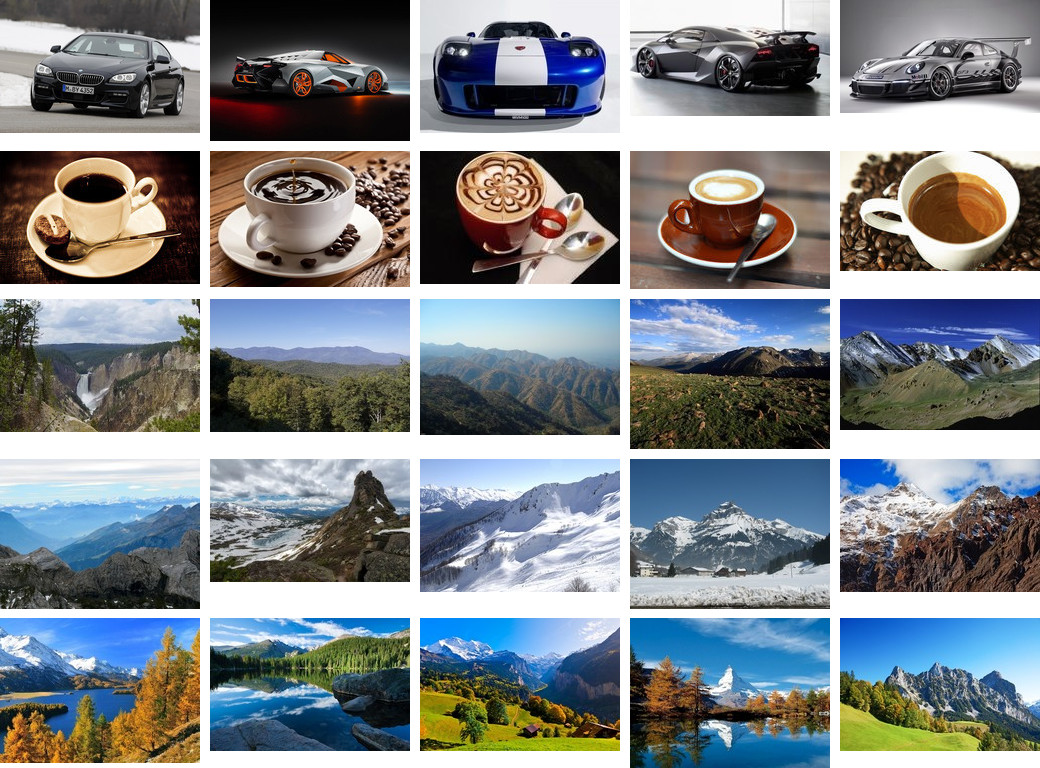}

    \caption{Results obtained when clustering the image dataset using
        APOC. The rows contain, car outliers, cup outliers, cluster 1,
        cluster 2 and cluster 3 respectively. The three clusters differ
        in their colour scheme: the first one has muted tones, the
        second is dominated by cold white and blue tones, while the third
        one contains vibrant colours.}
    \label{fig:image-data-results}
\end{figure}

\begin{table}[bt]
    \caption{Energy relative to the LP solution obtained on the natural
        scenes data set. APOC and LD obtain energy values which are
        nearly identical to LP and find all the correct outliers.
        \kmmm{} only finds suboptimal solutions and thus struggles to
        find good outliers unless the correct value for $k$ is
    specified.}
    \label{tab:lp-comparison}

    \centering
    \normalsize


    \begin{tabular}{lrrr}
        \toprule
        Method  & Energy    & Clusters  & Jaccard Index\\
        \midrule
        LP      & $100.0 \%$  & $3$       & $1$ \\
        APOC    & $96.6 \%$   & $3$       & $1$ \\
        LD      & $95.3 \%$   & $3$       & $1$ \\
        \kmmm   & $81.3 \%$   & $1$       & $0.54$ \\
        \kmmm   & $85.2 \%$   & $3$       & $0.93$ \\
        \kmmm   & $83.8 \%$   & $5$       & $0.64$ \\
        \bottomrule
    \end{tabular}
    \vspace{-5mm}
\end{table}

In this section we apply APOC to outlier detection in image collections.
Our dataset is composed of images from mountain ranges with outliers in
the form of cars and coffee cups. The distance between images is
computed from colour histograms and local binary pattern
\citep{ojala_2002} histograms using the Bhattacharyya distance.  When we
specify the correct number of outliers APOC and LD find all of the
images belonging to our two outlier groups and cluster the remaining
images according to their appearance. \figref{image-data-results} shows
outliers in the first two rows and examples belonging to the three
clusters found in subsequent rows as found by APOC. The three clusters
contain images which mainly differ in their colour and mood. The first
cluster contains images with muted greens and browns, the second cluster
has images with cold white and blue colours and finally the last cluster
is dominated by vibrant colours. For comparison we also process this
data using LP, LD and \kmmm{}. The results in \tabref{lp-comparison}
show how close the different methods come to the optimal energy, the
number of clusters and the quality of the selected outliers. This shows
that APOC and LD perform nearly as well as the optimal LP solution,
while \kmmm{} depends on a good choice of $k$ and still fails to recover
all the outliers. This experiment also highlights the capability of APOC
and LD to pick communities of points as outliers in a global scope.

\section{Related work}
\label{sec:related-work}

In \secref{introduction}, we provided some context for combining
clustering and outlier detection. Here we elaborate further on other
works which are germane to our problem. The problem of outlier detection
has been extensively researched in both statistics and data mining
\citep{chandola, hawkins}. However, both communities have different
perspectives. The statistical perspective is to design models which are
robust against outliers. In the process, outliers are discovered and
evaluated. The data mining perspective is to directly mine for outliers
and then determine their usefulness. The study of outlier detection in
data mining was pioneered by the work of \citet{knorr97} who proposed a
definition of distance-based outliers which relaxed strict
distributional assumptions and was readily generalisable to
multi-dimensional data sets. Following Knorr and Ng, several variations
and algorithms have been proposed to detect distance-based outliers
\citep{bay03, ramaswamy00}. However, the outliers detected by these
methods are global outliers, i.e., the ``outlierness'' is with respect
to the whole dataset.  \citet{breunig_2000} have argued that in some
situations local outliers are more important than global outliers and
cannot be easily detected by standard distance-based techniques.  They
introduced the concept of local outlier factor ($LOF$), which captures
how isolated an object is with respect to its surrounding neighbourhood.
The concept of local outliers has subsequently been extended in several
directions \citep{chandola,sunc06-2,papadimitriou}.

Despite the observation that clustering and outlier detection techniques
should naturally complement each other, research work in integrating the
two is limited. The MCD (as noted in the Introduction), is one prominent
approach where the problem of outliers was tightly integrated in an
optimisation framework \citep{rousseeuw_1999}. The \kmmm{}  approach
extended the idea of MCD to include clustering \citep{chawla_2013}. The
theoretical research in combining clustering and outlier detection has
been investigated in the context of robust versions of the k-median and
facility location problems \citep{chen_2008,charikar_2001}. However, as
noted before, the focus has exclusively been on deriving theoretical
approximation bounds. To the best of our knowledge the use of belief
propagation ideas to investigate outliers is entirely novel.

\section{Conclusion}
\label{sec:conclusion}

In this paper we presented a novel approach to joint clustering and
outlier detection formulated as an integer program. The proposed
optimisation problem enforces valid clusters as well as the selection of
a fixed number of outliers. We then described three ways of solving the
optimisation problem using (i) linear programming, (ii) affinity
propagation with outlier detection and (iii) Lagrangian duality.
Experiments on synthetic and real data show how the joint optimisation
outperforms two stage approaches such as \kmmm. Additionally,
experimental results suggest that results obtained via APOC and LD are
very close to the optimal solution at a fraction of the computational
time. Finally, we detail the modifications of the LD method, needed to
process large scale datasets.

\bibliographystyle{abbrvnat}
\bibliography{APOC}

\begin{thebibliography}{29}
\providecommand{\natexlab}[1]{#1}
\providecommand{\url}[1]{\texttt{#1}}
\expandafter\ifx\csname urlstyle\endcsname\relax
  \providecommand{\doi}[1]{doi: #1}\else
  \providecommand{\doi}{doi: \begingroup \urlstyle{rm}\Url}\fi

\bibitem[Arthur and Vassilvitskii(2007)]{arthur_2007}
D.~Arthur and S.~Vassilvitskii.
\newblock {k-means++: The Advantages of Careful Seeding}.
\newblock In \emph{ACM-SIAM Symposium on Discrete Algorithms}, 2007.

\bibitem[Bay and Schwabacher(2003)]{bay03}
S.~Bay and M.~Schwabacher.
\newblock {Mining distance-based outliers in near linear time with
  randomization and a simple pruning rule}.
\newblock In \emph{Int.~Conf.~on Knowledge Discovery and Data Mining}, 2003.

\bibitem[Bertsimas and Weismantel(2005)]{bertsimas_2005}
D.~Bertsimas and R.~Weismantel.
\newblock \emph{{Optimization over Integers}}.
\newblock Dynamic Ideas Belmont, 2005.

\bibitem[Breunig et~al.(2000)Breunig, Kriegel, Ng, and Sander]{breunig_2000}
M.~Breunig, H.~Kriegel, R.~Ng, and J.~Sander.
\newblock {LOF: Identifying Density-Based Local Outliers}.
\newblock In \emph{Int.~Conf.~on Management of Data}, 2000.

\bibitem[Chandola et~al.(2009)Chandola, Banerjee, and Kumar]{chandola}
V.~Chandola, A.~Banerjee, and V.~Kumar.
\newblock {Anomaly detection: A survey}.
\newblock \emph{ACM Computing Surveys}, 2009.

\bibitem[Charikar et~al.(2001)Charikar, Khuller, Mount, and
  Nara\-simhan]{charikar_2001}
M.~Charikar, S.~Khuller, D.~M. Mount, and G.~Nara\-simhan.
\newblock {Algorithms for Facility Location Problems with Outliers}.
\newblock In \emph{Proc.~of the ACM-SIAM Symposium on Discrete Algorithms},
  2001.

\bibitem[Chawla and Gionis(2013)]{chawla_2013}
S.~Chawla and A.~Gionis.
\newblock {k-means--: A Unified Approach to Clustering and Outlier Detection}.
\newblock In \emph{SIAM International Conference on Data Mining}, 2013.

\bibitem[Chawla and Sun(2006)]{sunc06-2}
S.~Chawla and P.~Sun.
\newblock {SLOM}: A new measure for local spatial outliers.
\newblock \emph{Knowledge and Information Systems}, 2006.

\bibitem[Chen(2008)]{chen_2008}
K.~Chen.
\newblock {A constant factor approximation algorithm for $k$-median clustering
  with outliers}.
\newblock In \emph{Proc.~of the ACM-SIAM Symposium on Discrete Algorithms},
  2008.

\bibitem[Croux and Ruiz-Gazen(1996)]{croux_1996}
C.~Croux and A.~Ruiz-Gazen.
\newblock {A Fast Algorithm for Robust Principal Components Based on Projection
  Pursuit}.
\newblock In \emph{Proc. in Computational Statistics}, 1996.

\bibitem[Cuesta-Albertos et~al.(1997)Cuesta-Albertos, Gordaliza, and
  Matr{\'a}n]{cuesta_1997}
J.~Cuesta-Albertos, A.~Gordaliza, and C.~Matr{\'a}n.
\newblock {Trimmed $k$-means: an attempt to robustify quantizers}.
\newblock \emph{The Annals of Statistics}, 1997.

\bibitem[Eiter and Mannila(1994)]{eiter_1994}
T.~Eiter and H.~Mannila.
\newblock {Computing Discrete Fr{\'e}chet Distance}.
\newblock Technical report, Technische Universit\"at Wien, 1994.

\bibitem[Frey and Dueck(2007)]{frey_2007}
B.~Frey and D.~Dueck.
\newblock {Clustering by Passing Messages Between Data Points}.
\newblock \emph{Science}, 2007.

\bibitem[García-Escudero et~al.(2010)García-Escudero, Gordaliza, Matrán, and
  Mayo-Iscar]{garcia_2010}
L.~García-Escudero, A.~Gordaliza, C.~Matrán, and A.~Mayo-Iscar.
\newblock {A Review of Robust Clustering Methods}.
\newblock \emph{Advances in Data Analysis Classification}, 2010.

\bibitem[Givoni and Frey(2009)]{givoni_2009}
I.~Givoni and B.~Frey.
\newblock {A Binary Variable Model for Affinity Propagation}.
\newblock \emph{Neural Computation}, 2009.

\bibitem[Hawkins(1980)]{hawkins}
D.~Hawkins.
\newblock \emph{{Identification of Outliers}}.
\newblock Chapman and Hall, London, 1980.

\bibitem[Hennig(2008)]{hennig_2008}
C.~Hennig.
\newblock {Dissolution point and isolation robustness: Robustness criteria for
  general cluster analysis methods}.
\newblock \emph{Journal of Multivariate Analysis}, 2008.

\bibitem[Huber and Ronchetti(2008)]{huber_2008}
P.~Huber and E.~Ronchetti.
\newblock \emph{{Robust Statistics}}.
\newblock Wiley, 2008.

\bibitem[Knorr and Ng(1997)]{knorr97}
E.~Knorr and R.~Ng.
\newblock {A Unified Notion of Outliers: Properties and Computation}.
\newblock In \emph{Int.~Conf.~on Knowledge Discovery and Data Mining}, 1997.

\bibitem[Komodakis et~al.(2008)Komodakis, Paragios, and
  Tziritas]{komodakis_2008}
N.~Komodakis, N.~Paragios, and G.~Tziritas.
\newblock {Clustering via LP-based Stabilities}.
\newblock In \emph{Advances in Neural Information Processing Systems}, 2008.

\bibitem[Kschischang et~al.(2001)Kschischang, Frey, and
  Loeliger]{kschischang_2001}
F.~R. Kschischang, B.~J. Frey, and H.-A. Loeliger.
\newblock {Factor Graphs and the Sum-Product Algorithm}.
\newblock \emph{IEEE Transactions on Information Theory}, 2001.

\bibitem[LeCun et~al.(1998)LeCun, Bottou, Bengio, and Haffner]{lecun_1998}
Y.~LeCun, L.~Bottou, Y.~Bengio, and P.~Haffner.
\newblock {Gradient-based learning applied to document recognition}.
\newblock \emph{Proceedings of the IEEE}, 1998.

\bibitem[Ojala et~al.(2002)Ojala, Pietikainen, and Maenpaa]{ojala_2002}
T.~Ojala, M.~Pietikainen, and T.~Maenpaa.
\newblock {Multiresolution Gray-Scale and Rotation Invariant Texture
  Classification with Local Binary Patterns}.
\newblock \emph{IEEE Transactions on Pattern Analysis and Machine
  Intelligence}, 2002.

\bibitem[Papadimitriou et~al.(2003)Papadimitriou, Kitagawa, Gibbons, and
  Faloutsos]{papadimitriou}
S.~Papadimitriou, H.~Kitagawa, P.~Gibbons, and C.~Faloutsos.
\newblock {LOCI}: Fast outlier detection using the local correlation integral.
\newblock In \emph{Int.~Conf.~on Data Engineering}, 2003.

\bibitem[Ramaswamy et~al.(2000)Ramaswamy, Rastogi, and Shim]{ramaswamy00}
S.~Ramaswamy, R.~Rastogi, and K.~Shim.
\newblock {Efficient Algorithms for Mining Outliers from Large Data Sets}.
\newblock In \emph{Int.~Conf.~on Management of Data}, 2000.

\bibitem[Rosenberg and Hirschberg(2007)]{rosenberg_2007}
A.~Rosenberg and J.~Hirschberg.
\newblock {V-Measure: A conditional entropy-based external cluster evaluation
  measure}.
\newblock In \emph{Proc.~of the Joint Conference on Empirical Methods in
  Natural Language Processing and Computational Natural Language Learning},
  2007.

\bibitem[Rousseeuw and Driessen(1999)]{rousseeuw_1999}
P.~Rousseeuw and K.~Driessen.
\newblock {A fast algorithm for the minimum covariance determinant estimator}.
\newblock \emph{Technometrics}, 1999.

\bibitem[Weiss et~al.(2007)Weiss, Yanover, and Meltzer]{weiss_2007}
Y.~Weiss, C.~Yanover, and T.~Meltzer.
\newblock {MAP Estimation, Linear Programming and Belief Propagation with
  Convex Free Energies}.
\newblock In \emph{Proc.~of the Conference on Uncertainty in Artificial
  Intelligence}, 2007.

\bibitem[Wright et~al.(2009)Wright, Ganesh, Rao, Peng, and Ma]{wright_2009}
J.~Wright, A.~Ganesh, S.~Rao, Y.~Peng, and Y.~Ma.
\newblock {Robust Principal Component Analysis: Exact Recovery of Corrupted
  Low-Rank Matrices by Convex Optimization}.
\newblock In \emph{Advances in Neural Information Processing Systems}, 2009.

\end{thebibliography}


\end{document}